\documentclass[conference]{IEEEtran}
\IEEEoverridecommandlockouts
\usepackage{cite}
\usepackage{amsmath,amssymb,amsfonts}
\usepackage{algorithmic}
\usepackage{graphicx}
\usepackage{textcomp}
\usepackage{tabularx}
\usepackage{booktabs}
\usepackage{xcolor}

\usepackage{multirow}
\usepackage{verbatim}

\def\BibTeX{{\rm B\kern-.05em{\sc i\kern-.025em b}\kern-.08em
    T\kern-.1667em\lower.7ex\hbox{E}\kern-.125emX}}
\begin{document}

\title{Impacts of Anthropomorphizing Large Language Models in Learning Environments}

\author{
\IEEEauthorblockN{Kristina Schaaff}
\IEEEauthorblockA{\textit{Dept. of IT \& Engineering} \\
IU International University of Applied Sciences, Germany \\
kristina.schaaff@iu.org}
\and 
\IEEEauthorblockN{Marc-André Heidelmann}
\IEEEauthorblockA{\textit{Dept. of Social Sciences} \\
IU International University of Applied Sciences, Germany\\
marc-andre.heidelmann@iu.org}
}

\maketitle


\section{Introduction}
Large Language Models (LLMs) are increasingly being used in learning environments to support teaching---be it as learning companions or as tutors~\cite{bahroun, ramadanis, Wollny2021}. With our contribution, we aim to discuss the implications of the anthropomorphization of LLMs in learning environments on educational theory to build a foundation for more effective learning outcomes and understand their emotional impact on learners.  

According to the media equation~\cite{reevesMediaEquationHow1996a}, people tend to respond to media in the same way as they would respond to another person. A study conducted by the Georgia Institute of Technology showed that chatbots can be successfully implemented in learning environments. In this study, learners in selected online courses were unable to distinguish the chatbot from a ``real'' teacher~\cite{kukulska-hulmeInnovatingPedagogyReport2021}. As LLM-based chatbots such as OpenAI's GPT series are increasingly used in educational tools, it is important to understand how the attribution processes to LLM-based chatbots in terms of anthropomorphization affect learners' emotions. 
\section{Problem Statement}
We know from learning research that learning and education are closely linked to emotions~\cite{schreyoeggEmotionenImCoaching2015}. Arnold even states that ``education is emotional maturity''~\cite{arnoldEmotionaleKonstruktionWirklichkeit2019}. In particular, negative emotional experiences such as irritation, limit experiences, or feelings of strangeness are given great relevance in qualitative educational research~\cite{kollerBildungAndersDenken2023}. 
In this context, the way learners perceive and interact with LLMs-based chatbots in educational environments can have a significant impact on educational experiences and outcomes. The anthropomorphization of these models, which attributes human-like characteristics to them, affects their integration and perception, thereby affecting their educational potential. In our research, we aim to explore the consequences of anthropomorphizing LLM-based chatbots in learning environments, focusing on user interaction, and learning effectiveness. In particular, educational theory and ethical considerations play a role.  

By supporting both--students and educators–-LLM-based chatbots are transforming the educational landscape. The emergence of LLM-based chatbots offers entirely new possibilities, as they are far more powerful than earlier chatbots~\cite{caldarini22} and are also able to behave empathically~\cite{schaaff2023exploring}.

\begin{figure}[htbp]
    \centering
    \includegraphics[width=0.9\linewidth]{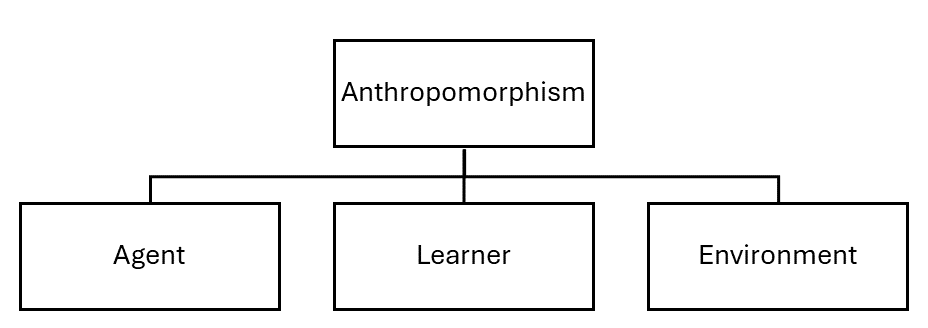}
    \caption{Factors of anthropomorphism in learning environments.}
    \label{fig:anthropomorphism}
\end{figure}

Similarly to the factors of anthropomorphism summarized by \cite{kim23}, we identified the following factors as relevant when LLM-based chatbots are used in learning scenarios: The learning \textit{agent}, i.e., chatbot, the \textit{learner} itself, and \textit{environmental} factors which influence the learner (see Figure~\ref{fig:anthropomorphism}).  

Looking at the \textit{agent}, several factors can contribute to anthropomorphization. Cognitive intelligence refers to the ability to perceive, reason, and act on problems; to combine efficient, useful, goal-oriented, and autonomous actions with effective output; and to produce and process natural language, imitate human cognitive functions, and mimic human interaction. Emotional intelligence refers primarily to the ability to perceive one's own and others' emotions and to communicate moods, emotions, and feelings~\cite{kim23, go19, troshani21, Moussawi2019}. Characteristics such as personality, in the sense of consistent behavior and adaptation of communication styles and preferences evoking human personality traits~\cite{alabed22}; personalization, in the sense of recognizing and responding to a learner's individual preferences, needs, and behaviors~\cite{sarraf24, alabed22}; and identity, which is created and shaped by a unique and recognizable character or brand, as well as its name, voice, appearance, and background story~\cite{alabed22, go19}, are also significant. Moreover, factors such as physical appearance, voice, movement, gestures, and facial expressions~\cite{kim23, sarraf24, go19} can influence anthropomorphism even though they are only relevant if an agent is accompanied by an avatar.

Regarding the \textit{learner}, there are several psychological determinants, such as emotions, motivation, and cognitive processes~\cite{kim23, kaplan19}, influencing the personality of a learner. The personality determines how a learner perceives an AI and interacts with it~\cite{kwak17, yang19, alabed22, kaplan19, epleySeeingHumanThreefactor2007}, and therefore its individual tendency to anthropomorphize technical systems~\cite{kim23}. Moreover, the individual tendency is influenced by self-congruence, i.e., the correspondence between the characteristics of an AI and the learner's self-image~\cite{alabed22, macinnis17, vandenhende14}.

Finally, sociological, and cultural studies highlight the relevance of macro-environmental factors as an important determinant of anthropomorphization. For example, shared values, beliefs, and practices are important when interacting with a learning agent. Moreover, cultural differences can significantly influence how AI systems are perceived and anthropomorphized \cite{kim23,epleySeeingHumanThreefactor2007}.

Several studies point to both, the positive and negative effects of anthropomorphizing chatbots for conducting learning processes. Anthropomorphism can lead to enhanced engagement and motivation among learners by providing a more relatable and interactive experience \cite{albrecht23}. Studies have shown that people tend to respond more positively to technology that exhibits human-like characteristics~\cite{nassAreMachinesGender1997}. \cite{faruk23} see a particularly positive aspect in overcoming learning challenges through anthropomorphic processes. 
However, excessive anthropomorphism can also set unrealistic expectations regarding the capabilities of LLMs, potentially leading to confusion or frustration~\cite{epleySeeingHumanThreefactor2007}. Moreover, \cite{holmes23} emphasize the risk of a lack of knowledge of reality and a fundamental dependence on technology. \cite{duggan16} also highlights that frustration can arise when systems do not meet human standards or are unable to respond appropriately to complex human questions or needs.
Therefore, the perception of LLM-based chatbots as `intelligent tutors' can influence the effectiveness of learning. Personalized feedback from anthropomorphized agents can enhance understanding and retention of information~\cite{mayerSocialCuesMultimedia2003}. However, the impact varies depending on the subject matter, the design of the agent's responses, and the learner's profile~\cite{soniImpactArtificialIntelligence2019}. 



\begin{figure}[htbp]
    \centering
    \includegraphics[width=0.9\linewidth]{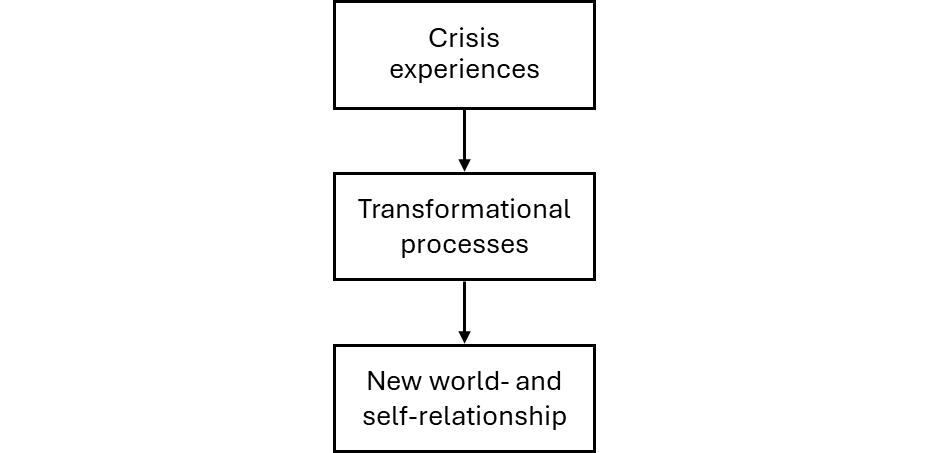}
    \caption{Theory of transformational education based on \cite{kollerBildungAndersDenken2023}.}
    \label{fig:transedu}
\end{figure}

The idea behind the theory of transformational education, which is influenced by biography theory, posits that learning is not only a linear process of collecting knowledge elements~\cite{heidelmannOrganisationenUndNetzwerke2022}. Instead, it is about changing how we understand things \cite{kollerBildungAndersDenken2023}. As illustrated in Figure \ref{fig:transedu}, this change can be triggered by \textit{crisis experiences} like irritation and strangeness \cite{kollerBildungAndersDenken2023}. These intense emotions are important for learning in general~\cite{schreyoeggEmotionenImCoaching2015}. 




When learning is triggered by crisis experiences this can lead to \textit{transformational processes}, disrupting the foundational frameworks that have structured an individual's life and guided their daily interpretations~\cite{kokemohr2007bildung}. This necessitates comprehensive educational processes that facilitate the development of a \textit{new world- and self-relationship}.

Consequently, \cite{marotzki1990entwurf} defines learning processes solely based on the change in the mode of information processing, regardless of the quality and nature of the information processed. Learning is understood as a 'transformation' \cite{marotzki1990entwurf}, in which the educational process does not take place within the existing orientation framework, but in the course of which it changes as a whole.

\section{Research Questions \& Methodology}
Learners use LLM-based chatbots to support their learning process. By using these systems, all learning activities can be tracked. This information can be connected with other learning materials that match the learner's level as a starting point for new learning goals. This focus on the learner and their integration into a ubiquitous, real-time, and opaque data structure could pose a problem in terms of educational theory.

From this perspective, the following questions arise: (1) Are LLM-based chatbots able to induce these intense emotions of irritation and strangeness when being anthropomorphized? If so, (2) do these emotions significantly influence the learning outcomes in learning environments using LLM-based chatbots? 

To evaluate these questions, we will set up a study based on the factors that contribute to the anthropomorphism of a system. For this study, we will develop two different learning systems: one system integrating the relevant factors of anthropomorphism and one which does not. We will implement a decision-making task which allows us to capture the performances as well as the decision-making times of the participants. 
The two systems will be analyzed in a comparative study with a large cohort of students from IU International University of Applied Sciences. Furthermore, we will evaluate the emotional states of the participants during the task using questionnaires. 

\section{Summary \& Conclusion}

As LLMs continue to evolve, their anthropomorphization will likely play a crucial role in their acceptance and utility in educational contexts. Future research should focus on optimizing the balance between relatability and realism in LLM interactions, developing guidelines for their use, and exploring innovative applications in personalized learning.
The anthropomorphization of LLM-based chatbots in learning environments presents both: Opportunities and challenges. While it can enhance engagement and learning effectiveness, it also raises ethical concerns and the potential for negative impacts on user experience, including unrealistic expectations and emotional discomfort. 

In educational science, it is assumed that strong emotions contribute to the initiation of educational processes in learners. Especially for learning with LLM-based chatbots, the question of the effect of emotions on individual learning is a desideratum. In our study, we plan to investigate whether and to what extent the anthropomorphization of AI-based systems can evoke such emotions. As educationalists and engineers, we consider both the implications of educational theory and the technical implementation and control options. This interdisciplinary approach addresses the highly relevant desideratum of technical-pedagogical development and processing of AI-supported education at universities. Our findings can help educators create more effective educational technologies by creating a better understanding of the balance between making AI relatable and maintaining realistic expectations of its capabilities.

\bibliographystyle{IEEEtran}
\bibliography{10_bibliography}

\end{document}